# Content-Conditioned Generation of Stylized Free-hand Sketches

Jiajun Liu
*Xidian University*
Xi'an, China
jiajunliu@stu.xidian.edu.cn

Siyuan Wang
*Xidian University*
Xi'an, China
siyuanwang@stu.xidian.edu.cn

Guangming Zhu
*Xidian University*
Xi'an, China
gmzhu@xidian.edu.cn

Liang Zhang
*Xidian University*
Xi'an, China
liangzhang@xidian.edu.cn

Ning Li
*China Institute of Marine Technology & Economy*
Beijing, China
lining199008@163.com

Eryang Gao
*China Institute of Marine Technology & Economy*
Beijing, China
thugaoeryang@163.com

*Abstract*—In recent years, the recognition of free-hand sketches has remained a popular task. However, in some special fields such as the military field, free-hand sketches are difficult to sample on a large scale. Common data augmentation and image generation techniques are difficult to produce images with various free-hand sketching styles. Therefore, the recognition and segmentation tasks in related fields are limited. In this paper, we propose a novel adversarial generative network that can accurately generate realistic free-hand sketches with various styles. We explore the performance of the model, including using styles randomly sampled from a prior normal distribution to generate images with various free-hand sketching styles, disentangling the painters' styles from known free-hand sketches to generate images with specific styles, and generating images of unknown classes that are not in the training set. We further demonstrate with qualitative and quantitative evaluations our advantages in visual quality, content accuracy, and style imitation on SketchIME.

## I. Introduction

In the military field, the situation map can help commanders better understand the battlefield situation, so as to formulate corresponding combat plans. As a key element of the situation map, the military symbols can provide an effective way to identify military units and entities, and provide rich information for the situation map. Among many situational mapping systems, the hand-drawn sketch-based situational mapping system has the advantages of being more efficient and intuitive. However, the key of such systems is to recognize the user's free-hand sketching military symbols. In recognition tasks, the use of neural networks has always been a mainstream solution. The recognition model based on neural network needs rich data to improve the accuracy. However, due to the secrecy of the military, it is impossible to carry out large-scale free-hand sketches sampling. Different people have different free-hand sketching styles[1], if the training data is sampled from a few people, the model's generalization ability will become worse. Thus, to expand the dataset, using a generative adversarial network (GAN)[2] is a good solution to generate free-hand sketches with various styles. This task is challenging with GAN, since it faces the following three difficulties:

- A wealth of information on free-hand sketches. The free-hand sketches are composed of components[1]. The number, position and type of the components will affect the class of sketches. When generating a free-hand sketches, it is necessary to keep the components and the content stable.

- Various styles. Just like handwriting, people will have different free-hand sketching styles when drawing the same sketch class. It is difficult to have good generalization ability of the trained classification model by only sampling the data from a small number of people. Therefore, it is necessary to generate free-hand sketches with various styles.

- Difficulty imitating style. Most free-hand sketches lack rich texture information, and the geometric structure may change when the same person draws the same sketch class multiple times[1].

To address the above problems, we propose a free-hand sketching imitation GAN. The proposed model can generate images with various free-hand sketching styles. It can also disentangle free-hand sketching styles from reference samples to flexibly control the style of the generated sketches. Our contributions are summarized as follows:

- We propose a novel free-hand sketches generator.

- Ability to generate free-hand sketches not only the categories within training set, but also the categories outside of it.

- Ability to generate free-hand sketches with various styles and also ability to transfer disentangled free-hand sketching styles to other categories.

## II. Related Work

### A. Sketch Generation

Sketch-related generation tasks are always a hot topic, David Ha et al. (2017) proposed SketchRNN[3] for generating serialized sketches, which adjusts the model parameters by reducing the error between the real sketch sequence and the generated sketch sequence so that it is as close as possible to the real image. Image-to-image transformation models, such as Pix2Pix(Phillip Isola et al.)[4] and CycleGAN (Jun-Yan Zhu 2017)[5], can also be utilized to generate images from different domains. Runtao Liu et al. (2019) [6]proposed to generate realistic shoe images using free-hand sketches. Although its purpose is to generate realistic shoe images, it can also generate corresponding sketches from real shoe images because it is based on CycleGAN. Further, Yong-Jin Liu et al. (2020)[7] utilized an asymmetric cyclic map to transform photos of people into three styles of sketches by incorporating style features into the generator. However, it can

This work was supported by the Chinese Defense Advance Research Program (50912020105).

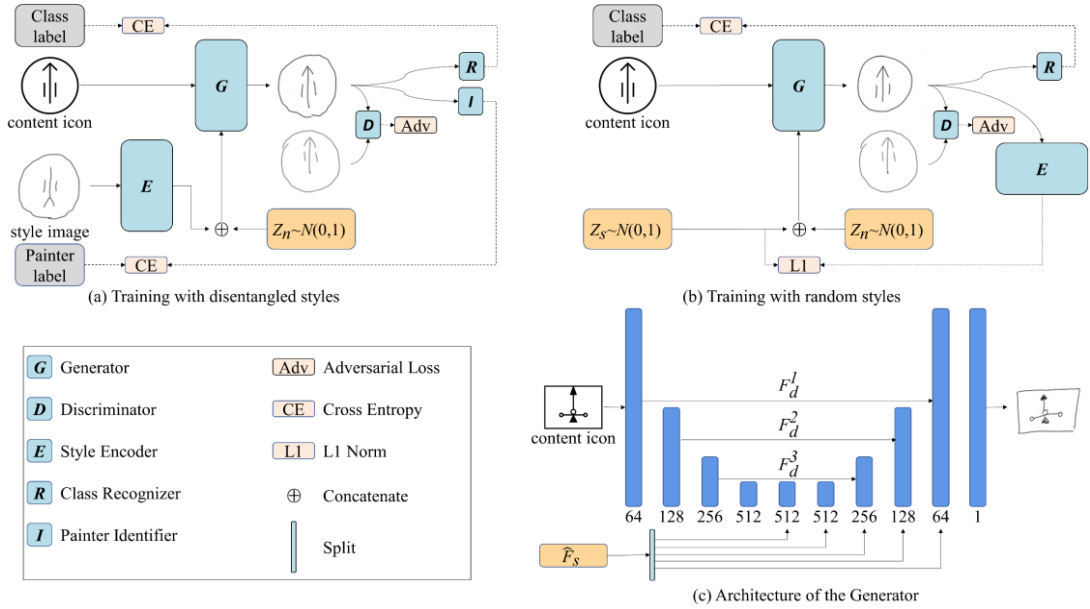

Fig. 1. Architecture of the proposed content-conditioned generation model.

only generate sketches of these three fixed styles, and the style is not consistent with the free-hand sketching style.

*B. Neural Style Transfer*

A subclass of image-to-image translation is style transfer, which recombines disentangled content elements and style elements to generate images with different styles while maintaining their underlying content[8]. Gatys et al. (2016) proposed Neural Style Transfer (NST) [8], which uses a pre-trained convolutional neural network to capture the content information and the style information of images and generate new images using the content images and the style images. It has a profound impact on the research of image style transfer. In Im2Pencil (Li et al. 2019) [9], the styles of the generated images can be controlled, such as shading and sketchiness, enabling a more personalized and controllable sketch generation process. However, the styles described above and the styles of the free-hand sketches are not consistent. The previous style information mainly exists in the texture, whereas the free-hand sketching style information mainly exists the lines and the geometric structure. Moreover, the styles of the free-hand sketches are much more varied, and individual samples containing much less style information[1].

### III. METHODOLOGY

Our goal is to generate realistic free-hand sketches with various styles, so there are three problems need to be solved. Firstly, the free-hand sketch lacks texture information. Category supervision and self-supervision are introduced to control the generated content. Secondly, the free-hand sketches require a variety of styles. Style information is injected into the generator to influence the free-hand sketching styles of generated images. Thirdly, free-hand sketches with a specific style need to be generated. The specific style is disentangled and injected into the generator, and then the style of the generated image is supervised. In this section, we describe how to design our method to achieve them.

*A. Problem Formulation*

For the convenience of description, let $SD = \{X,Y,M,P\}$ be a free-hand sketches dataset, and let $d_c = (x, y, m, p\,) \in SD$ be a content sample, where $x$ is the real free-hand sketch of the sample, i.e., the real image, $y$ is the class label, $m$ is the content icon corresponding to the class label, and $p$ is the author of the sample, i.e. the painter label. Since the free-hand sketching style needs to be disentangled from the style image, a sample $d_s$ is also needed as a style sample. Let $d_s = (x_s, y_s, m_s, p_s) \in SD, x_s \in X$, where $x_s$ is a real free-hand sketch, which is used as a style reference image here, i.e., the style image.

For each content sample $d_c$, there will be a corresponding style sample $d_s$, and the $d_s$ satisfies the rules of random selection, that is

$$d_s = \{(x_s, y_s, m_s, p_s) | Random(SD)\}. \quad (1)$$

The task is divided into two parts, one of which is the ability to generate realistic and diverse free-hand sketch $\bar{x}$ using generator $G$ that is

$$\bar{x} = G(m, z_s), \quad (2)$$

where $z_s$ is random style features, and the other is the ability to generate free-hand sketch $\bar{x}$ with specific styles, that is

$$\bar{x} = G(m, F_s), \quad (3)$$

where $F_s$ are reference style features. Specifically, an overview of our model is given in Fig. 1(a) & (b).

*B. Network Architecture*

The proposed architecture is composed of a conditional generator $G$, a discriminator $D$, a style encoder $E$, a painter identifier $I$ and a class recognizer $R$.

**Generator.** In the whole network structure, the function of the generator $G$ is to generate realistic free-hand sketches with various styles, the details are as follows:

- Content acquisition. Using the standardized symbols as content icon $m$ can provide the geometric

information to the generator $G$. Specifically, the general architecture of the generator follows the U-Net[10] architecture with skip connections, as the Fig. 1(c). In the encoding stage, the content icon $m$ is encoded into a high-dimensional latent space through three times of down-sampling. The feature maps $F_d^j$ containing geometric information is obtained for each layer. In the decoding stage, in order to gradually recover lost geometric information, the feature maps $F^{i-1}$ are gradually enlarged and concatenated with the feature maps $F_d^j$ from the corresponding layer of the encoding stage in the channel dimension. $F^i$ can be implemented as:

$$F^i = concat(Up(F^{i-1}), F_d^j), j \in \{1, 2, 3\}, \quad (4)$$

where $Up$ is up-sampling and $concat$ is concatenation.

- Random style acquisition. To make the generated images with diverse style, we use $z_s$ randomly sampled from a prior normal distribution as random style features. There is unconscious variation when the same person draws the same sketch class multiple times. Therefore, to mimic this phenomenon, additive noise $z_n \sim N(0,1)$ is applied to the output latent space to obtain slightly distorted style features $\widehat{F}_s$, that is

$$\widehat{F}_s = concat(z_s, z_n). \quad (5)$$

- Reference style acquisition. To make the generated images with a specific style, the free-hand sketching style is disentangled from the style image $x_s$ as a reference style $F_s$ using style encoder $E$, that is

$$F_s = E(x_s). \quad (6)$$

We also use the additive noise $z_n$ to obtain the style features $\widehat{F}_s$, that is

$$\widehat{F}_s = concat(F_s, Z_n). \quad (7)$$

The acquired $\widehat{F}_s$ is divided into equal chunks and conditional batch normalization (CBN) [11] is used to inject the style chunks into the generator $G$, which affects the free-hand style of the resulting image (such as the line slant, geometric structure, and filled-in texture, etc.).

**Discriminator**. The discriminator $D$ is mainly used for binary classification to determine whether the image input to the discriminator is a real image $x$ from the training set or a fake image $\bar{x}$ generated by the generator $G$.

**Style Encoder**. The style encoder $E$ is used to disentangle the style features from the reference image, which from the style image $x_s$ or the generated fake image $\bar{x}$. Consequently, the style encoder $E$ does not need to explicitly access the identifier of the corresponding painter.

**Painter Identifier**. Painter Identifier $I$ is used to identify which painter the current image belongs to. The role of the Painter Identifier $I$ is to supervise the style encoder $E$, guide the style encoder to cluster embedding for samples from the same painter, and scatter embedding for samples from different painters. It also supervises the generator $G$, which makes the style of the generated image $\bar{x}$ closer to the style image $x_s$.

**Recognizer**. Recognizer $R$ is similar to Painter Identifier $I$. It is used to judge which class the current image belongs to. The role of the Recognizer is to supervise the generated content by the generator $G$, so that the geometric structure of the generated image $\bar{x}$ are more in line with the corresponding class.

*C. Objective Functions*

**Adversarial Loss**. The generator $G$ can not only use the random style $z_s$ to generate images with various styles, but also the style of generated images close to the reference images, therefore the adversarial loss $L_{adv}$ is divided into two parts, i.e.,

$$L_{adv} = L_{adv1} + L_{adv2}. \quad (8)$$

For any content icon $m$ (standardized symbol) and a random style feature $z_s$, the generator $G$ learns to generate a fake image through the adversarial loss, so that the discriminator $D$ can't distinguish between generated images and real images, that is

$$L_{adv1} = E_x[\log D(x)] + E_{m, z_s}\left[\log\left(1 - D(G(m, z_s))\right)\right]. \quad (9)$$

In addition, when using a given style image $x_s$ as a reference sample, the generator $G$ needs to generate a specific image based on the disentangled style by the style encoder $E$, that is

$$L_{adv2} = E_x[\log D(x)] + E_{m, x_s}\left[\log\left(1 - D(G(m, E(x_s)))\right)\right]. \quad (10)$$

**Class Recognition Loss**. The generator $G$ needs to use the content icon $m$ to generate an image containing the content of the corresponding class. Therefore, the class recognizer $R$ needs to be introduced to supervise the generator $G$. The recognizer is optimized by minimizing the cross-entropy loss for each ground-truth pair $\{x, y\}$ (where $x \in X, y \in Y$), i.e.,

$$L_{class}^D = E_{x, y}[-y \log R(x)]. \quad (11)$$

When minimizing the adversarial loss, $R$ keeps fixing its parameters, and the recognizer $R$ uses the $L_{class}^G$ as in (12) to force the generator $G$ to generate realistic free-hand sketches, and to ensure that the content of the corresponding class.

$$L_{class}^G = E_{y, m, x_s}\left[-y \log R\left(G(m, E(x_s))\right)\right] + \\ E_{y, m, z_s}\left[-y \log R(G(m, z_s))\right] \quad (12)$$

$L_1$ **Loss**. When using a random style $z_s$ to generate an image, the model is forced to reconstruct the style of the generated image, by minimizing the $L_1$ loss of the random style and the reconstructed style, to ensure that the random style and the reconstructed style are close. That is the random style has a guiding effect, so as to generate images with various styles, i.e.,

$$L_{style} = E_{m, z_s}\left[\|z_s - E(G(m, z_s))\|_1\right]. \quad (13)$$

**Painter Identifier Loss**. Being able to generate images that are stylistically consistent with the style image $x_s$ is one of our goals. The style features $F_s$ of free-hand sketches drawn by the same painter should be close, and the style features of different painter should vary widely. So the style encoder $E$ need to be optimized. We need to ensure that the generator $G$ can obtain the semantic information of the style features $\widehat{F}_s$ to generate the images conforming to a specific style. Therefore, a painter identifier $I$ is needed to constrain the style encoder $E$ and generator $G$. The painter identifier is optimized by minimizing the cross-entropy loss for each ground-truth pair $\{x, p\}$ (where $x \in X$, $p \in P$), to ensure that the identifier $I$ correctly predicts the painter, i.e.,

$$L_{id}^D = E_{x,p}[-p \log I(x)]. \quad (14)$$

When minimizing the adversarial loss, $I$ fixing its parameters. In order to better learn disentangled style features $F_s$ from style images $x_s$, the generated image $\bar{x}$ and the style image $x_s$ are forced to have highly similar styles. This is achieved by minimizing the loss function, i.e.,

$$L_{id}^G = E_{x_s, m, p_s}\left[-p_s \log I\big(G(m, E(x_s))\big)\right]. \quad (15)$$

**KL Loss**. Finally, the latent space encoded by the style encoder $E$ is further regularized to match a priori normal distribution, as

$$L_{kl} = E_{x_s}[D_{KL}(E(x_s) || N(0,1))], \quad (16)$$

thus ensuring the quality of the generated images using the random style, where $D_{KL}$ denotes the Kullback-Leibler Divergence (*KL*-divergence).

The complete objective function can be summarized as follows:

- When maximizing the adversarial loss, the discriminator $D$, the class recognizer $R$, and the painter identifier $I$ are respectively optimized as

$$L_D = -L_{adv}, \quad L_R = L_{class}^D, \quad L_I = L_{id}^D \quad (17)$$

- When minimizing the adversarial loss, the generator $G$ and the style encoder $E$ are jointly optimized as

$$L_{G,E} = L_{adv} + \lambda_{class} L_{class}^G + \lambda_{id} L_{id}^G + \lambda_{style} L_{style} + \lambda_{kl} L_{kl}, \quad (18)$$

where $\lambda$ represents the importance of each loss function.

## IV. EXPERIMENTS

### A. Dataset.

To evaluate our model, we use the latest free-hand sketches dataset SketchIME[12], which consists of 374 categories, and each painter only draws part of the classes.

Considering the style differences between free-hand sketching and handwriting, most of the classes containing characters were eliminated in our experiments. Specifically, 341 classes are used, and the training set and test set are divided according to the rule that the painters are mutually exclusive. In each class, 16 images of each painter are selected, if the painter exists.

### B. Baseline

In the past, the image style transfer was mainly texture-based, and line-style transfer was mainly focused on handwritten characters, numbers, and text. However, the transfer of free-hand sketching styles to sketch generation has not been fully explored. Therefore, in our experiments, we can only compare our model with the handwritten text generation model GANwriting (Lei Kang 2020) [13].

Specifically, the SketchIME is converted into the dataset format required by GANwriting. At the same time, the text encoding and text decoding are respectively converted to class encoding and class decoding. Each generated image is no longer a multi-character text image, but a single-class free-hand sketch. Especially, the height and width of all generated images are 128 pixels in our experiments.

### C. Implementation Details.

Our experiments are all trained on a single NVIDIA GeForce RTX 3090 GPU with 24 GB. Our model is trained for 69 epochs, and the batch size is set to 8. During model training, the model is optimized using Adam algorithm, where the initial learning rate is 0.0001, $(\beta_1, \beta_2) = (0.5, 0.999)$, and the learning rate started to decay linearly at the 25th epoch. Moreover, we make $\lambda_{kl}$ equal to 0.001, and the remaining $\lambda$ are dynamically adjusted using a gradient balancing strategy.

When training GANwriting, the epochs are set to 15000, and the default settings for others are used directly if available.

### D. Evaluation Metrics.

We evaluate GANs using comprehensive metrics. The first is commonly used metrics to evaluate the visual quality of generated images, including Fréchet Inception Distance (*FID*), Kernel Inception Distance (*KID*), Inception Score (*IS*), and Peak Signal-to-Noise Ratio (*PSNR*), where *FID* and *KID* are measures of similarity between the generated image and the real image, *IS* is a measure of the quality and diversity of the generated image, and *PSNR* is used to measure the reconstruction error of the generated image.

In addition to the conventional evaluation metrics for generative model, we also train another class recognizer $\widetilde{R}$ and painter identifier $\widetilde{I}$ using an independent dataset to measure the content and style of the generated images. This dataset also from the SketchIME, and samples from 34 painters. Especially these samples don't appear in the training set and the test set of our model. Let the correct rate of class recognition is C_ACC, and the correct rate of painter recognition is P_ACC.

## V. EXPERIMENTAL RESULTS

### A. Random-Guided Generation.

The random style $z_s$ is randomly sampled from a prior normal distribution. Using the random style $z_s$ as a style guide, free-hand sketches can be generated free-hand sketches with various styles.

Specifically, as shown in the Fig. 2, the first row is the content icons, and then each of the next rows is generated images according to the same style, and each column is images of the same class generated using different random styles.

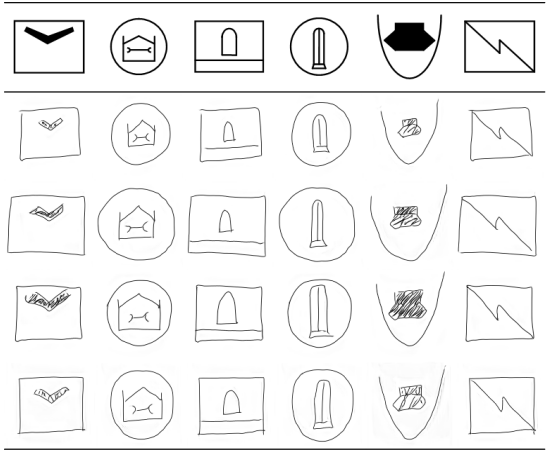

Fig. 2. Random-guided generation. Random styles of generated images are sampled from a prior normal distribution.

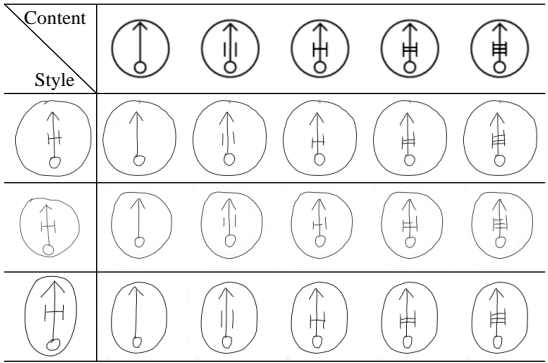

Fig. 3. Reference-guided generation. Different styles of generated images are disentangled from reference images.

The images generated by different random styles appear rich variations in line slant, filled-in texture, and geometry. Using the same random style as a guide, the styles of generated images are similar, even though they are from different classes.

*B. Reference-Guided Generation.*

In order to imitate the painter's style to generate some samples of classes they have not drawn. Our model utilizes the reference style $F_s$ disentangled from the style image $x_s$ to generate free-hand sketches with a specific style.

As shown in Fig. 3, it can be seen that our model successfully imitates the free-hand sketching styles similar to the style images, even if those painters have not painted these classes. The geometrical structures of the images generated using the same styles from similar classes are similar. It is proving that our style encoder $E$ can disentangle the reference style $F_s$ from style image, and our generator can correctly identify the semantic information of the reference style $F_s$, that is, the disentangled style can guide the generation of the image.

*C. Style Interpolation.*

To further explore the style space of the model, two widely differing free-hand sketching styles are disentangled in some classes, and then interpolation styles are obtained using multiple linear interpolation.

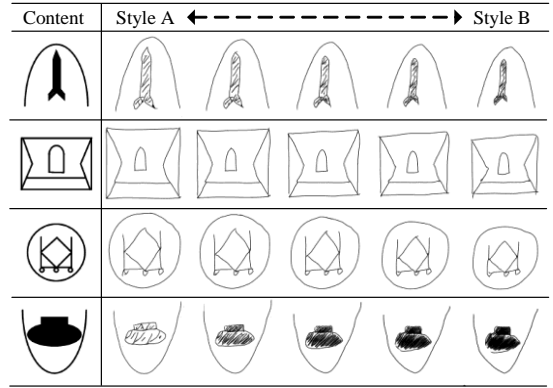

Fig. 4. Images generated using interpolation styles between two different styles.

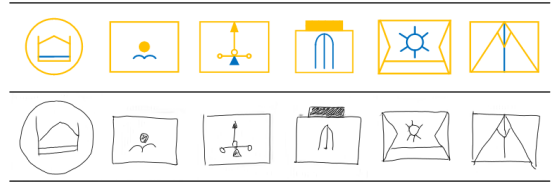

Fig. 5. Generated image results for option 1. The blue components and the orange components appear separately in the other training classes.

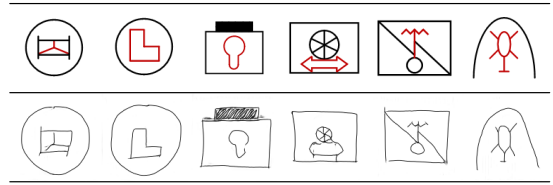

Fig. 6. Generated image results for option 2. The red components never appear in the training set.

The result is shown in Fig. 4, where these interpolation styles gradually guide generated images from style A to style B, while ensuring that the contents of the generated images are stabilized.

The result confirms the guiding function of style. Moreover, the gradual change of the generated images proves that the style space of the model is continuous, that is, our model can generalize in the style space, instead of just remembering the trivial visual information of the painters in the training set.

*D. Class Extension*

To verify the ability of the generator in class extension, i.e., generating classes that do not exist in the training set. The dataset SketchIME is re-divided, and 256 classes are selected from it to participate in model training. Additionally, 50 classes and 15 classes are respectively selected for testing. The painters of the test set are present in the training set, and the classes of the referenced style images are not equal to the classes of the generated images.

Option 1, the test classes are a combination of components from other training classes. As shown in Fig. 5, there are 50 test classes of this type.

Option 2, the components of the test classes are independent of the training classes. As shown in Fig. 6, there are 15 test classes of this type.

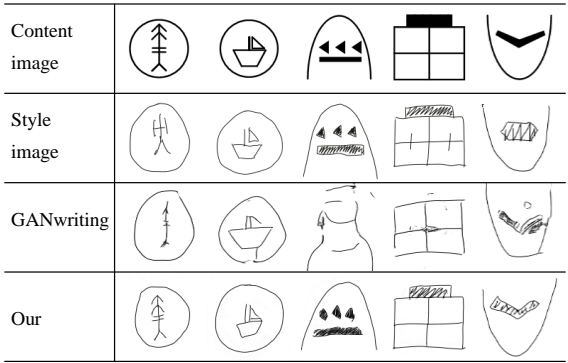

Fig. 7. Qualitative comparison of different GANs.

TABLE I. THE ACCURACY OF THE TEST CLASS IN THE CLASS EXTENSION

| Test Set | C_ACC | | | P_ACC | |
|---|---|---|---|---|---|
| | Real | Noise | Style | Real | Style |
| Option 1 | 99.638% | 97.613% | 96.625% | 97.763% | 96.913% |
| Option 2 | 99.833% | 98.125% | 96.083% | 97.583% | 91.708% |

[a.] C_ACC is the accuracy rate of the class; P_ACC is the accuracy rate of the painter;
[b.] Real is real free-hand images; Noise and Style are the random style and reference style.

TABLE II. QUANTITATIVE COMPARISON OF DIFFERENT GANS

| Model | FID ↓ | KID ↓ | IS ↑ | PSNR ↑ |
|---|---|---|---|---|
| GANwriting | 23.548 | 1.086 | 2.625 | 14.118 |
| Our | 12.175 | 0.353 | 2.924 | 14.247 |

TABLE III. ACCURACY COMPARISON OF DIFFERENT GANS

| Model | C_ACC ↑ | P_ACC ↑ |
|---|---|---|
| GANwriting | 73.311% | 22.060% |
| Our | 99.976% | 39.783% |
| Ground Truth | 99.706% | 97.064 |

The result is shown in Fig. 5&Fig. 6, from the perspective of visual quality, the generated image details of Option 1 are more complete compared to Option 2, since the classes of Option 1 consist of components that have been seen by the generator. These test classes are further evaluated by the $\tilde{R}$ and the $\tilde{I}$, and the results are shown in TABLE I. C_ACC shows that the generated images can be as close as possible to real free-hand sketches for unseen classes. P_ACC indicates that the generator can transfer the free-hand sketching styles of painter in the training data well to unseen classes.

*E. Compared with the baseline.*

To demonstrate the superiority of our model, we make qualitative and quantitative comparisons of the competing GAN.

For qualitative comparisons, during GANwriting training, each generated image needs 50 style images to disentangle the style features. In order to enable qualitative comparisons, GANwriting is kept consistent with our model during testing, that is using only one style image per generated image. Specifically, as shown in Fig. 7, the visual quality of images generated by our model is significantly better than the GANwriting.

For quantitative comparisons, comprehensive evaluation metrics are used. The specific results are shown in TABLE II and TABLE III. Our model is better than GANwriting in the quality and diversity of generated images and the similarity between generated images and real images. The results of C_ACC and P_ACC show that our model is better than GANwriting in image content generation and style imitation.

The shortcoming of our model is that the images generated using the style images do not exactly match the images drawn by the painter. This is because only one style image is used for each image generated, and a single image contains less style information. Overall, our model outperforms the GANwriting model on these metrics. Experiments demonstrate the superiority of our model in free-hand sketches generation.

## VI. CONCLUSIONE

In this paper, we propose a novel free-hand sketches generation model. It can generate realistic free-hand sketches with diverse styles using random styles randomly sampled from a prior normal distribution. It is also possible to use reference styles disentangled from the style images to generate free-hand sketches with specific styles. Even if the content icons are outside the training, our model is able to roughly generate the corresponding free-hand sketches. Qualitative and quantitative experimental comparisons verify that our model outperforms the compared GAN in content accuracy, style migration ability, and visual quality.

REFERENCES

[1] P. Xu, T. M. Hospedales, Q. Yin, Y. -Z. Song, T. Xiang and L. Wang, "Deep Learning for Free-Hand Sketch: A Survey," in IEEE Transactions on Pattern Analysis and Machine Intelligence, vol. 45, no. 1, pp. 285-312, 1 Jan. 2023, doi: 10.1109/TPAMI.2022.3148853.

[2] Goodfellow, J. Pouget-Abadie, M. Mirza, B. Xu, D. Warde-Farley, S. Ozair, A. Courville, and Y. Bengio, "Generative adversarial nets," in Proceedings of the Neural Information Processing Systems Conference, 2014.

[3] D. Ha and D. Eck, "A neural representation of sketches," arXiv preprint arXiv:1704.03477, 2017.

[4] P. Isola, J.-Y. Zhu, T. Zhou, and A. A. Efros, "Image-to-Image Translation with Conditional Adversarial Networks," in Proceedings of IEEE International Conference on Computer Vision (ICCV), 2017, pp. 5967–5976.

[5] Z. X. Zhu, J. Y. Park, P. Isola, and A. A. Efros, "Unpaired Image-to-Image Translation Using Cycle-Consistent Adversarial Networks," in Proceedings of the IEEE International Conference on Computer Vision (ICCV), 2017, pp. 2223–2232.

[6] Runtao Liu, Qian Yu, and Stella Yu, "Unsupervised Sketch-to-Photo Synthesis," arXiv preprint arXiv: 1909.08313, 2019.

[7] R. Yi, Y.-J. Liu, Y.-K. Lai, and P. L. Rosin, "Unpaired Portrait Drawing Generation via Asymmetric Cycle Mapping," in Proceedings of IEEE Conference on Computer Vision and Pattern Recognition (CVPR), 2020, pp. 8214-8222.

[8] L. A. Gatys, A. S. Ecker, and M. Bethge, "Image Style Transfer Using Convolutional Neural Networks," in IEEE Conference on Computer Vision and Pattern Recognition (CVPR), 2016, pp. 2414-2423.

[9] Y. Li, C. Fang, A. Hertzmann, E. Shechtman, and M.-H. Yang, "Im2Pencil: Controllable Pencil Illustration from Photographs," in Proceedings of the IEEE Conference on Computer Vision and Pattern Recognition (CVPR), 2019.

[10] O. Ronneberger, P. Fischer, and T. Brox, "U-net: Convolutional networks for biomedical image segmentation," in Medical Image Computing and Computer-Assisted Intervention–MICCAI 2015: 18th International Conference, Munich, Germany, October 5-9, 2015, Proceedings, Part III 18, Springer International Publishing, 2015, pp. 234-241.

[11] H. De Vries, F. Strub, J. Mary, et al., "Modulating early visual processing by language," in Advances in Neural Information Processing Systems, vol. 30, 2017.

[12] G. Zhu, S. Wang, Q. Cheng, K. Wu, H. Li, and L. Zhang, "Sketch InpuMethod Editor: A Comprehensive Dataset and Methodology for Systematic Input Recognition," in Proceedings of ACM Multimedia (MM), 2023.

[13] L. Kang, P. Rib, Y. M. R.-A. F. M. V., "GANwriting: Content-Conditioned Generation of Styled Handwritten Word Images," in Proceedings of the European Conference on Computer Vision (ECCV), 2020.